\pdfoutput=1

\documentclass[11pt]{article}

\usepackage[preprint]{acl}

\usepackage{times}
\usepackage{latexsym}
\usepackage{amssymb}
\usepackage{amsmath}
\usepackage{multirow}
\usepackage{booktabs}
\usepackage{mathrsfs}
\usepackage{pifont}

\newcommand{\cmark}{\ding{51}}%
\newcommand{\xmark}{\ding{55}}%
\usepackage[T1]{fontenc}

\usepackage[utf8]{inputenc}

\usepackage{microtype}

\usepackage{inconsolata}

\usepackage{graphicx}

\title{UniTabNet: Bridging Vision and Language Models for Enhanced \\
	Table Structure Recognition}

\author{
  \textbf{Zhenrong Zhang\textsuperscript{1}\footnotemark[1]},
  \textbf{Shuhang Liu\textsuperscript{1}\footnotemark[1]},
  \textbf{Pengfei Hu\textsuperscript{1}},
  \textbf{Jiefeng Ma\textsuperscript{1}}, 
\\
  \textbf{Jun Du\textsuperscript{1}\footnotemark[2]},
  \textbf{Jianshu Zhang\textsuperscript{2}},
  \textbf{Yu Hu\textsuperscript{2}}
\\
  \textsuperscript{1}University of Science and Technology of China,
  \textsuperscript{2}iFLYTEK AI Research,
}

\begin{document}
\maketitle
\begin{abstract}
In the digital era, table structure recognition technology is a critical tool for processing and analyzing large volumes of tabular data. Previous methods primarily focus on visual aspects of table structure recovery but often fail to effectively comprehend the textual semantics within tables, particularly for descriptive textual cells. In this paper, we introduce UniTabNet, a novel framework for table structure parsing based on the image-to-text model. UniTabNet employs a ``divide-and-conquer'' strategy, utilizing an image-to-text model to decouple table cells and integrating both physical and logical decoders to reconstruct the complete table structure.  We further enhance our framework with the Vision Guider, which directs the model's focus towards pertinent areas, thereby boosting prediction accuracy. Additionally, we introduce the Language Guider to refine the model's capability to understand textual semantics in table images. Evaluated on prominent table structure datasets such as PubTabNet, PubTables1M, WTW, and iFLYTAB, UniTabNet achieves a new state-of-the-art performance, demonstrating the efficacy of our approach. The code will also be made publicly available.
\end{abstract}

\renewcommand{\thefootnote}{\fnsymbol{footnote}} 
\footnotetext[1]{Equal contribution.}
\footnotetext[2]{Correspondence author. \href{mailto:jundu@ustc.edu.cn}{jundu@ustc.edu.cn}}

\section{Introduction}

In this era of knowledge and information, documents serve as crucial repositories for various cognitive processes, including the creation of knowledge databases, optical character recognition (OCR), and document retrieval. Among the various document elements, tabular structures are particularly notable. These structures distill complex information into a concise format, playing a pivotal role in fields such as finance, administration, research, and archival management~\citep{SurveyTable}. Table structure recognition (TSR) focuses on converting these tabular structures into machine-readable data, facilitating their interpretation and utilization. Therefore, TSR as a precursor to contextual document understanding will be beneficial in a wide range of applications~\citep{DeCNT,DeepDeSRT}.

\begin{figure}[t]
	\includegraphics[width=\columnwidth]{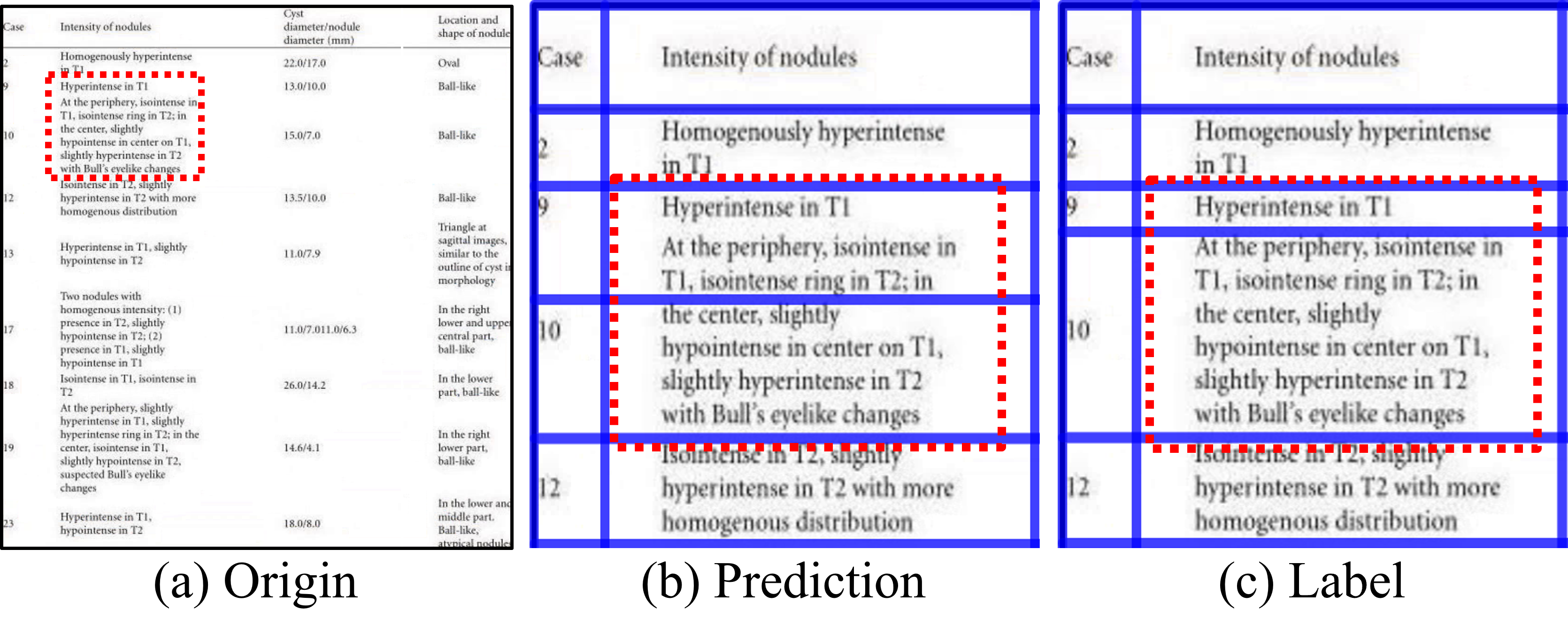}
	\caption{The illustration of the rich textual features in tabular images. (a) displays the original tabular image. (b) and (c) provide zoomed-in views of the area outlined by the red dashed box in (a). (b) shows the prediction result of the recent state-of-the-art table structure recognition method SEMv2\citep{SEMv2}. (c) presents the ground truth label for table structure. The red dashed box highlights the discrepancy between the prediction and the ground truth label.}
	\label{fig:01_problem}
\end{figure}

Table images efficiently convey information through visual clues, layout structures, and plain text. However, most previous methods\citep{SciTSR, WTW, SEMv2} in TSR primarily utilize visual or spatial features, neglecting the textual content within each table cell. The structures of some tables exhibit inherent ambiguities when assessed solely based on visual appearance, especially for wireless tables which contain cells with descriptive content, as illustrated in Figure~\ref{fig:01_problem}. To enhance accuracy in TSR, it is crucial to leverage the cross-modality characteristics of visually-rich table images by jointly modeling both visual and textual information~\citep{ERNIE-Layout}.

Recent advancements in document understanding, exemplified by methods such as Donut~\citep{Donut} and Pix2Struct~\citep{Pix2Struct}, have embraced an end-to-end image-to-text paradigm. These approaches leverage the Transformer architecture~\citep{Transformers} during pre-training to decode OCR results, demonstrating superior perception of image content. By diminishing reliance on traditional OCR engines, they exhibit remarkable adaptability across diverse document understanding tasks, highlighting their robust ability to comprehend text embedded in images. Despite these advancements, the application of this framework to TSR remains unexplored. While there are related works \citep{TableFormer,VAST} that employ this framework, they primarily focus on reconstructing table structures from a visual perspective, without adequately addressing the depth of textual understanding in images.

In this work, we adopt the image-to-text framework and introduce a visually linguistic unified model for TSR, named UniTabNet. This model is built on a ``divide and conquer'' design philosophy, initially using the image-to-text model to decouple table cells. According to the attributes of the table structure~\citep{SurveyTable}, the decoupled cells contain two types of attributes: logical and physical. The logical attributes cover the row and column span information of each cell, while the physical attributes include the bounding box coordinates of the cells. To parse these attributes independently, we design a logical decoder and a physical decoder. Since table images differ significantly from regular document images, each step of the decoding output is grounded in a clear visual basis, specifically visual cues from rows, columns, and cells. Therefore, we design a Vision Guider module, which directs the model to focus on relevant areas and make more precise predictions. Furthermore, to enhance the UniTabNet's understanding of text content in images, we develop a Language Guider. This module enables the model to perceive the corresponding text content at each decoding step, thereby understanding the textual semantics within the image. Experimental results on multiple public TSR datasets, such as PubTables1M~\citep{PubTables1M}, PubTabNet~\citep{PubTabNet}, iFLYTAB~\citep{SEMv2}, and WTW~\cite{WTW}, demonstrate that our approach achieves state-of-the-art performance, validating the effectiveness of our method.
The main contributions of this paper are as follows:
\begin{itemize} 
	\item We introduce UniTabNet, a unified visually linguistic model for TSR that adheres the ``divide and conquer'' strategy by first separating table cells, then using both logical and physical decoders to reconstruct the table structure.
	
	\item We develop the Vision Guider module, designed to direct the model's focus towards critical areas such as rows and columns, thereby enhancing the overall prediction accuracy.
		
	\item We enhance UniTabNet with the Language Guider module, which enhances the model's ability to perceive textual content within images, thereby improving its accuracy in predicting the structure of tables rich in descriptive content.
	
	\item Based on our proposed method, we achieve state-of-the-art performance on publicly available datasets such as PubTabNet, PubTables1M, WTW and iFLYTAB.
\end{itemize}

\section{Related Work}
Due to the rapid development of deep learning in documents, many deep learning-based TSR approaches have been presented. These methods can be roughly divided into three categories: bottom-up methods, split-and-merge based methods and image-to-text based methods.

One group of bottom-up methods~\citep{SciTSR,Res2tim,NCGM} treat words or cell contents as nodes in a graph and use graph neural networks to predict whether each sampled node pair belongs to the same cell, row, or column. These methods depend on the availability of bounding boxes for words or cell contents as additional inputs, which are challenging to obtain directly from table images. To eliminate this assumption, another group of methods~\citep{TabStructNet,LGPMA} has proposed directly detecting the bounding boxes of table cells. After cell detection, they design some rules to cluster cells into rows and columns. However, these methods regard the cells as bounding box, which is difficult to handle the cells in distorted tables. Other methods~\citep{LORE,WTW} detect cells through detecting the corner points of cells, making them more suitable for handling distorted cells. Nevertheless, they suffer from tables containing a lot of empty cells and wireless tables.

Split-and-merge based methods initially split a table into basic grid pattern, followed by a merging process to reconstruct the table cells.
Previous methods~\citep{SPLERGE,SEMv1} utilize semantic segmentation~\citep{FCN} for identifying rows, columns within tables in the ``split'' stage. However, segmenting table row/column separation lines in a pixel-wise manner is inaccurate due to the limited receptive field, and heuristic mask-to-line modules designed with strong assumptions in split stage make these methods work only on tables in digital documents. To enhance the accuracy of grid splitting in distorted tables, RobustTabNet~\citep{RobustTabNet} uses a spatial CNN-based separation line predictor to propagate contextual information across the entire table image in both horizontal and vertical directions. SEMv2~\citep{SEMv2} formulates the table separation line detection as the instance segmentation task. The table separation line can be accurately obtained by processing the table separation line mask in a row-wise/column-wise manner. TSRFormer with SepRETR~\citep{TSRFormerv1} formulates the table separation line prediction as a line regression problem and regresses separation line by DETR~\citep{Detr}, but it can't regress too long separation line well. TSRFormer with DQ-DETR~\citep{TSRFormerv2} progressively regresses separation lines, which further enhances localization accuracy for distorted tables.

Image-to-text based methods conceptualize the structure of tables as sequential data (HTML or LaTeX), utilizing an end-to-end image-to-text paradigm to decode table structures. The EDD model~\cite{PubTabNet} employs an encoder-dual-decoder architecture to generate both the logical structure and the cell content. During the decoding phase, EDD utilizes two attention-based recurrent neural networks; one is tasked with decoding the structural code of the table, while the other decodes the content. Building on this framework, TableFormer~\cite{TableFormer} employs a transformer-based decoder to enhance the capabilities of EDD's decoder. Additionally, it introduces a regression decoder that predicts bounding boxes rather than content, thus refining the focus on spatial elements. Addressing the challenge of limited local visual cues, VAST \cite{VAST} redefines bounding box prediction as a coordinate sequence generation task and incorporates a visual alignment loss to achieve more accurate bounding box outcomes.

\begin{figure}[t]
	\includegraphics[width=1\linewidth]{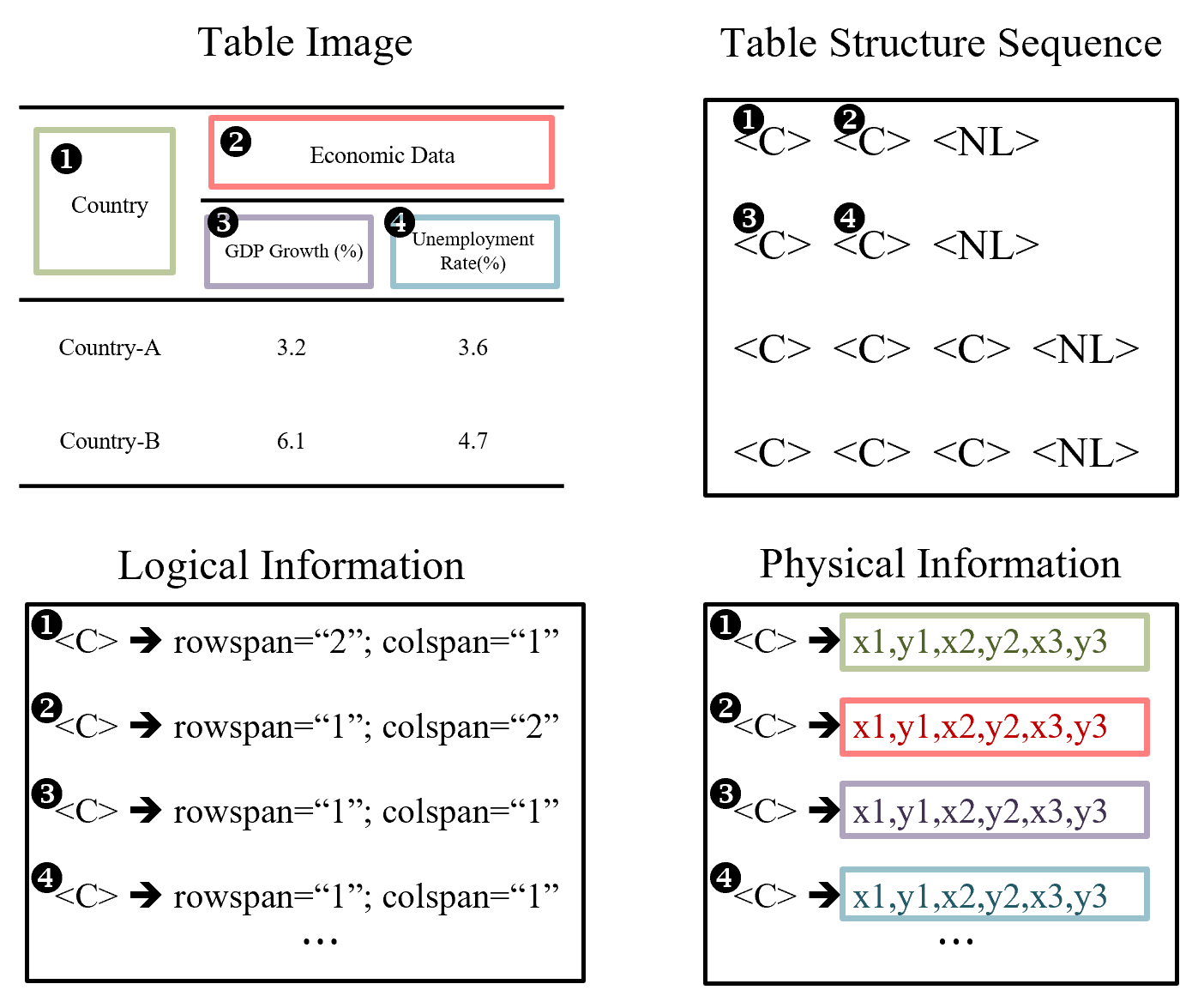}
	\caption{The illustration of the table structure recognition task.}
	\label{fig:02_taskdefinition}
\end{figure}

\section{Task Definition}
As illustrated in Figure~\ref{fig:02_taskdefinition}, given a table image $\boldsymbol{I}\in \mathbb{R}^{H\times W\times 3}$, our objective is to enable the model to predict the table structure sequence $\boldsymbol{S}=\left\{ \boldsymbol{s}_{{i}}\in \mathbb{R}^v \mid i=1,\ldots,T \right\}$, where $T$ is the length of the sequence and $v$ is the the size of token vocabulary, to reconstruct the table's layout. Previous methods~\citep{EDD,TableFormer,VAST} have employed various formats for the output table structure sequence $\boldsymbol{S}$, such as HTML and LaTeX. In contrast, our approach simplifies the decoding process significantly by using only two types of tokens: <C> and <NL>. <C> denotes a table cell, and <NL> indicates a newline, facilitating a concise representation of the table structure. According to the attributes of the table structure~\citep{SurveyTable}, each table cell encompasses both logical attribute $\boldsymbol{l} = \{ l_\text{row}, l_\text{col} \mid l_\text{row}, l_\text{col} \in \mathbb{N}^+ \}$ and physical attribute $\boldsymbol{p}=\{ p_j\in \mathbb{N} \mid j=1,\ldots , 8\}$. The logical attribute $\boldsymbol{l}$ specifies the cell's span across rows and columns, while the physical attribute $\boldsymbol{p}$ defines the spatial positioning of the cell within the image. Consequently, the output of our proposed model, UniTabNet, includes the structure sequence $\boldsymbol{S}$, along with logical attributes  $\boldsymbol{L}=\left\{ \boldsymbol{l}_{{i}}\in \mathbb{R}^2 \mid i=1,\ldots,T \right\}$ and physical attributes $\boldsymbol{P}=\left\{ \boldsymbol{p}_{{i}}\in \mathbb{R}^8 \mid i=1,\ldots,T \right\}$, providing a comprehensive description of the table’s layout.

\begin{figure*}[t]
	\includegraphics[width=1\linewidth]{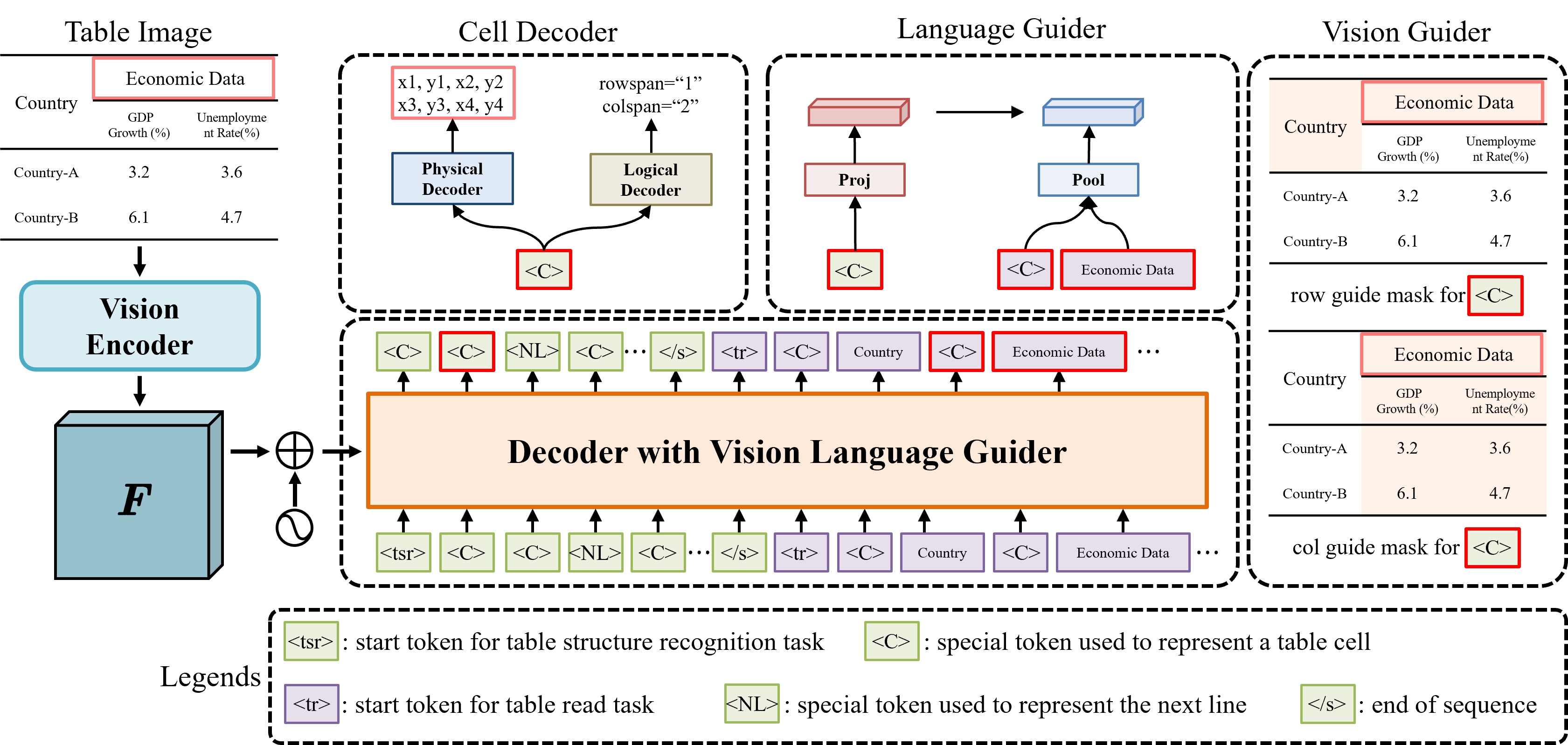}
	\caption{The overall architecture of UniTabNet. It mainly consists of a vision encoder and a text decoder. Using the text decoder's output, the Cell Decoder decodes the physical and logical attributes of table cells. The Vision Guider directs the model's focus on row and column information, while the Language Guider aids in understanding textual semantics.}
	\label{fig:03_unitabnet}
\end{figure*}

\section{Methodology}
As illustrated in Figure~\ref{fig:03_unitabnet}, UniTabNet is built upon the Donut~\citep{Donut} and primarily consists of a vision encoder and a text decoder, which decodes image features to generate the table structure sequence $\boldsymbol{S}$. To further decode the logical and physical attributes contained within each cell, we additionally design a logical decoder and a physical decoder to predict the cell attributes $\boldsymbol{l}$ and $\boldsymbol{p}$, respectively. Considering the nature of table images, we incorporate a Vision Guider and a Language Guider at the output of the text decoder. The Vision Guider directs the model to focus on relevant areas during cell decoding, while the Language Guider aids in understanding the corresponding textual information within the cells. Detailed descriptions of these modules will follow.

\noindent\textbf{Vision Encoder.} The vision encoder converts the table image $\boldsymbol{I}$ into a set of embeddings $\boldsymbol{Z} = \{\boldsymbol{z}_i \in \mathbb{R}^D \mid i = 1, \ldots, N\}$, where $N$ is feature map size and $D$ is the dimension of the latent vectors of the encoder. As depicted in Figure~\ref{fig:03_unitabnet}, we adopt the Swin Transformer~\citep{SwinT} as our primary vision backbone, following the Donut, to encode $\boldsymbol{I}$ into feature map $\boldsymbol{F}$. Additionally, we incorporate positional encoding~\citep{Transformers} into $\boldsymbol{F}$ to generate the final vision embeddings $\boldsymbol{Z}$.

\noindent\textbf{Text Decoder.} Similar to Donut, we utilize the BART~\citep{BART} decoder to generate the table structure sequence $\boldsymbol{S}$, conditioned on the $\boldsymbol{Z}$. Since UniTabNet is trained to predict the next tokens like LLMs~\citep{GPT4}, it only requires maximizing the likelihood of loss at training time.
\begin{align}
	\mathscr{L}_\text{lm}=\max \sum_{i=1}^T{\log P\left( \boldsymbol{s}_i\left| \boldsymbol{Z}, \boldsymbol{s}_{1:i} \right. \right)} 
\end{align}

\noindent\textbf{Physical Decoder.}
Given the output $\boldsymbol{H} = \{ \boldsymbol{h}_i \in \mathbb{R}^D \mid i = 1, \ldots, T \}$ from the last layer of the text decoder, the physical decoder decodes these hidden states to obtain the polygon coordinates $\boldsymbol{p}_i$ in the image. To facilitate this prediction, we introduce a set of 1,000 special tokens—<0>, <1>, ..., <999>—which are utilized for quantizing the coordinates of the polygons, forming a specialized vocabulary $\boldsymbol{Loc} \in \mathbb{R}^{1000 \times D}$. Specifically, for each coordinate point $p_j$ in the polygon $\boldsymbol{p}_i$, the prediction process is as follows: The corresponding hidden state $\boldsymbol{h}_i$ is transformed via a linear mapping to produce the $\boldsymbol{h}_{i}^{p_j}$, which serves as a query against the vocabulary $\boldsymbol{Loc}$. Unlike previous method~\citep{Pix2seq}, which perform direct classification over the location vocabulary, we define the final position of $p_j$ as the expected location based on the distribution given by $\boldsymbol{h}_{i}^{p_j}$ over $\boldsymbol{Loc}$:
\begin{align}
	\boldsymbol{h}_{i}^{p_j}=\text{Linear}\left( \boldsymbol{h}_i \right)
\end{align}
\begin{align}
	\boldsymbol{a}^{p_j}=\text{softmax} \left( \boldsymbol{h}_{i}^{p_j}\boldsymbol{Loc}^\top \right) 
\end{align}
\begin{align}
	E\left( p_j \right) =\sum_{i=0}^{999}{i\cdot a_{i}^{p_j}}
\end{align}
The polygon regression loss is defined as follows:
\begin{align}
	\mathscr{L}_\text{poly}=\frac{1}{8} \sum_{j=1}^8 \left(E(p_j) - p_j^*\right)^2
\end{align}
where $p_j^*$ denotes the ground truth label.

\noindent\textbf{Logical Decoder.}
The logical decoder predicts the rowspan and colspan information $\boldsymbol{L}$ for table cells based on the output $\boldsymbol{H}$ from the final hidden state of the text decoder. To illustrate, for predicting the rowspan information $l_\text{row}$ within $\boldsymbol{l}_i$, the hidden state $\boldsymbol{h}_i$ is first mapped through a matrix transformation to a vector $\boldsymbol{h}_{i}^{l_\text{row}}$. The $\boldsymbol{h}_{i}^{l_\text{row}}$ then serves as a query, computing the dot product with entries in the vocabulary $\boldsymbol{Loc}$, resulting in a score vector $\boldsymbol{a}^{l_\text{row}}$. The rowspan information $l_\text{row}$ is then determined by locating the index of the maximum value in the score vector $\boldsymbol{a}^{l_\text{row}}$.
\begin{align}
	\boldsymbol{h}_{i}^{l_\text{row}}=\text{Linear}\left( \boldsymbol{h}_i \right) 
\end{align}
\begin{align}
	\boldsymbol{a}^{l_\text{row}}=\boldsymbol{h}_{i}^{l_\text{row}}\boldsymbol{Loc}^\top
\end{align}
\begin{align}
	l_\text{row}=\text{argmax} \left( \boldsymbol{a}^{l_\text{row}} \right) 
\end{align}
Given the extreme imbalance in the distribution of rowspan and colspan across cells, we optimize our model using sigmoid focal loss~\citep{focalloss}. The span prediction loss for the logical decoder is defined as follows:
\begin{align}
	\mathscr{L}_\text{span}=L_f\left( \boldsymbol{a}^{l_\text{row}},\boldsymbol{l}_\text{row}^{*} \right) +L_f\left( \boldsymbol{a}^{l_\text{col}},\boldsymbol{l}_\text{col}^{*} \right) 
\end{align}
where $L_f$ represents the sigmoid focal loss function. The vectors $\boldsymbol{l}_\text{row}^{*}$ and $\boldsymbol{l}_\text{col}^{*}$ are one-hot representations of the ground truth span information for rowspan and colspan, respectively.

\noindent\textbf{Vision Guider.}
Unlike conventional document images, table images exhibit significant interdependencies among cells within the same row, column, or cell block. To enhance the model's ability to accurately capture these details during the decoding process, we develop the Vision Guider. This mechanism enables the model to focus more on the row and column information for each cell during decoding. Specifically, to capture the same row visual cues, we input the last layer's output $\boldsymbol{h}_i$ of the decoder into a matrix mapping to generate vector $\boldsymbol{h}_i^\text{row}$. The vector $\boldsymbol{h}_i^\text{row}$, serving as the query, is then used to fetch attention scores $\boldsymbol{a}^\text{row}$ from the visual embedding $\boldsymbol{Z}\in \mathbb{R}^{N\times D}$. A similar approach is adopted for the same column information $\boldsymbol{a}^\text{col}$.
\begin{align}
	\boldsymbol{h}_{i}^\text{row}=\text{Linear}\left( \boldsymbol{h}_i \right) 
\end{align}
\begin{align}
	\boldsymbol{a}^\text{row}=\boldsymbol{h}_{i}^\text{row}\boldsymbol{Z}^\top
\end{align}
The loss function for the Vision Guider is defined as:
\begin{align}
	\mathscr{L}_\text{vis}=L_f\left( \boldsymbol{a}^\text{row},\boldsymbol{g}_\text{row}^{*} \right) +L_f\left( \boldsymbol{a}^\text{col},\boldsymbol{g}_\text{col}^{*} \right) 
\end{align}
where $L_f$ denotes the sigmoid focal loss function, and $\boldsymbol{g}_\text{row}^{*}$ and $\boldsymbol{g}_\text{col}^{*}$ represent the row and column mask maps, respectively.

\noindent\textbf{Language Guider.}
Tables present data relationships in an exceedingly concise format. Beyond the prevalent numerical tables, there are also descriptive table images. To accurately recognize these descriptive tables, it is imperative that the model comprehends the content within the table to make more precise structural predictions. To this end, we introduce the Language Guider, which directs the model to understand the textual semantic information in the table. As illustrated in Figure~\ref{fig:04_taskdesign}, during the training phase, in addition to the essential Table Structure Recognition (TSR) task, we design an additional task named Table Read (TR), which prompts the model to sequentially output the content within table images, thereby enhancing the model's understanding of the text in the images. To ensure that the tokens in TSR possess text comprehension abilities similar to those in TR, we align the tokens from both tasks. Specifically, suppose a token <C> in TSR produces an output $\boldsymbol{h}_i$ at the decoder's last layer; we first map  $\boldsymbol{h}_i$ to $\boldsymbol{h}^\text{lang}_i$ using a matrix mapping. The corresponding token for <C> in TR, represented as $\boldsymbol{h}_{\left[n:m \right]}$ at the decoder's last layer, is then subject to mean pooling to produce $\boldsymbol{h}_\text{lang}^{*}$. Subsequently, a mean squared-error (MSE) loss is applied between $\boldsymbol{h}^\text{lang}_i$ and $\boldsymbol{h}_\text{lang}^{*}$, thus endowing TSR tokens with substantial text perception capabilities.
\begin{align}
	\boldsymbol{h}_{i}^\text{lang}=\text{Linear}\left( \boldsymbol{h}_i \right)
\end{align}
\begin{align}
	\boldsymbol{h}_\text{lang}^{*}=\text{Mean}\left( \boldsymbol{h}_{\left[n:m \right]} \right) 
\end{align}
\begin{align}
	\mathscr{L}_\text{lang}=\text{MSE}\left(\boldsymbol{h}_{i}^\text{lang},\boldsymbol{h}_\text{lang}^{*} \right) 
\end{align}

\section{Implementation Details}
Our methodology employs the following hyperparameters: The longest side of the image is resized to 1600 while maintaining the original aspect ratio. The downsampling factor of the visual backbone is set to 32. The dimension $D$ of the feature is set to 1024. The decoders consist of a stack of 4 identical layers, and the number of multi-heads is set to 16.

\begin{figure}[t]
	\includegraphics[width=1\linewidth]{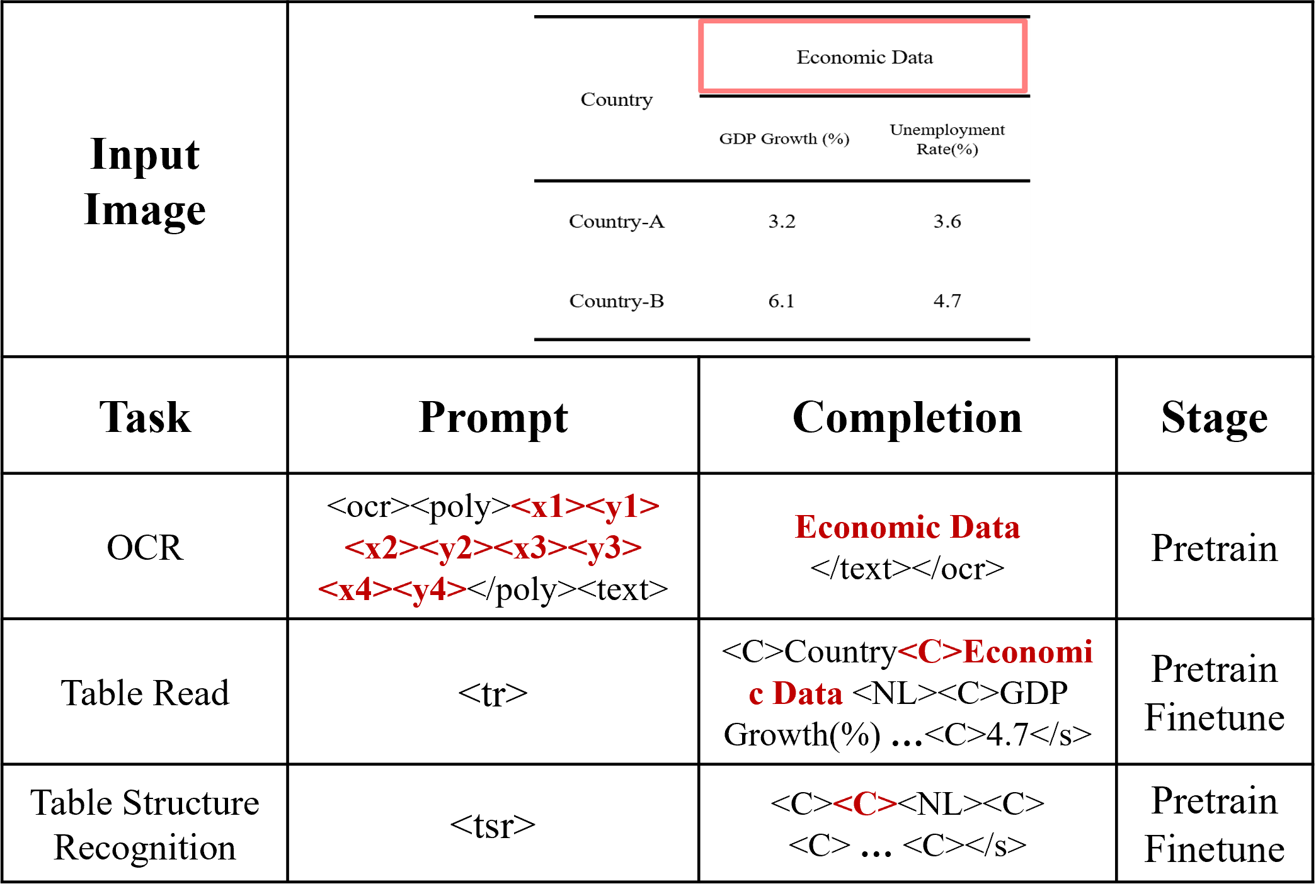}
	\caption{The illustration of the task design.}
	\label{fig:04_taskdesign}
\end{figure}

\noindent\textbf{Training.}
To train UniTabNet, we design three training tasks as depicted in Figure~\ref{fig:04_taskdesign}. These tasks aim to enable the model to comprehensively perceive tabular images. Specifically, the training process is divided into two phases. Initially, during the pre-training phase, we use a synthetic dataset comprising 1.4 million Chinese and English entries from SynthDog~\citep{Donut}, along with the training set from PubTables1M~\citep{PubTables1M}. After pre-training, the model is fine-tuned on specialized datasets dedicated to table structure recognition. 
We fine-tune UniTabNet using the Adam~\citep{Adam} optimizer with the learning rate of $5\times10^{-5}$.The learning rate is linearly warmed up over the first 10\% steps then linearly decayed. The training is conducted on 8 Telsa A40 48GB GPUs. The model is trained for 100 epochs on the iFLYTAB~\citep{SEMv2} and WTW~\citep{WTW} datasets, and for 10 epochs on the PubTables1M and PubTabNet~\citep{EDD} datasets.

In the overall loss of UniTabNet, there are primarily two categories: regression losses ($\mathscr{L}_\text{poly}$, $\mathscr{L}_\text{lang}$) and classification losses ($\mathscr{L}_\text{lm}$, $\mathscr{L}_\text{span}$, $\mathscr{L}_\text{vis}$). Given the significant scale differences among these losses, it is necessary to adjust their coefficients. Inspired by~\citep{WeightLoss} , we optimize the model by maximising the Gaussian likelihood with homoscedastic uncertainty.
\begin{align}
	\label{eq:ul_loss}
	\mathscr{L}_\text{total}=\sum_{k=1}^5{\frac{1}{2\sigma _{k}^{2}}\mathscr{L}_k+\log \left( 1+\sigma _{k}^{2} \right)}
\end{align}
The $\sigma$ is a learnable factor that adaptively adjusts the weight ratios among these losses. $\mathscr{L}_k$ represents the five losses mentioned above.

\noindent\textbf{Inference.}
During the inference phase, we feed the <tsr> token into UniTabNet and utilize a greedy search algorithm to decode the table structure sequence $\boldsymbol{S}$. Relying on the hidden states $\boldsymbol{H}$ from the last layer of the decoder, we can decode the physical $\boldsymbol{P}$ and logical $\boldsymbol{L}$ information corresponding to each cell. This allows for the complete reconstruction of the table structure.

\section{Experiments}
\subsection{Datasets and Evaluation Metrics}
To fully demonstrate the effectiveness of the UniTabNet, we conduct experiments across four datasets. Firstly, for single-scene electronic document table images, we select two representative datasets, PubTabNet~\citep{EDD} and PubTables1M~\citep{PubTables1M}, for evaluation. We assess these datasets using the TEDS-Struct~\citep{EDD} and GriTS~\citep{Grits} metrics to ensure comprehensive and comparative results. For complex scene table images, we chose the WTW~\citep{WTW} and iFLYTAB~\citep{SEMv2} datasets for evaluation, employing the F1-Measure~\citep{F1-Measure} and TEDS-Struct metrics to quantify the model's performance. Notably, we also extract a subset from the iFLYTAB validation set, termed iFLYTAB-DP, which comprises 322 descriptive table images. For more details on the datasets and evaluation metrics, please refer to the Appendix~\ref{app:dataset_and_metric}.

\begin{table}[t]
	\centering
	\caption{Comparison on PubTables1M}
	\label{tab:comparison_pubtables1m}
	\resizebox{\columnwidth}{!}{%
	\begin{tabular}{@{}llcc@{}}
		\toprule
		{Type}          & {Method} & {GriTS-Top} & {GriTS-Loc}  \\ \midrule
		\multirow{2}{*}{\parbox{1.8cm}{Bottom-up}} 
		& {Faster RCNN} & 86.16 & 72.11        \\
		& {DETR}        & 98.45  & \textbf{97.81}            \\ \midrule
		\multirow{2}{*}{\parbox{1.8cm}{Image-to-Text}} 
		& {VAST}        & 99.22  & 94.99            \\
		& Ours          & \textbf{99.43} & {95.37}   \\ \bottomrule
	\end{tabular}
}
\end{table}

\begin{table*}[t]
\centering
\caption{Comparison with SOTA methods across different datasets. \textbf{Bold} indicates the best result.}
\label{tab:comparison_with_sota}
\begin{tabular}{p{2cm}@{}p{4.4cm}cccccc@{}}
	\toprule
	{Type} & {Method} & {PubTabNet} & \multicolumn{3}{c}{{WTW}} & {iFLYTAB} \\ 
	\cmidrule(lr){3-3} \cmidrule(lr){4-6} \cmidrule(lr){7-7}
	&                & TEDS-Struct       & P     & R     & F1    & TEDS-Struct \\ 
	\midrule
	\multirow{3}{*}{\parbox{2cm}{Bottom-up}} & Cycle-CenterNet~\citep{WTW} & -             & 93.3  & 91.5  & 92.4  & -           \\
	& LORE~\citep{LORE}            & -             & 94.5  & \textbf{95.9} & \textbf{95.1} & -           \\
	& LGPMA~\citep{LGPMA}           & 96.70         & -     & -     & -     & -           \\
	\midrule
	\multirow{6}{*}{\parbox{2cm}{Split-and-merge}} 
	& SEM~\citep{SEMv1}             & 96.30         & -     & -     & -     & {75.9}        \\
	& RobustTabNet~\citep{RobustTabNet} & 97.00         & -     & -     & -     & -           \\
	& TSRFormer~\citep{TSRFormerv1}     & \textbf{97.50}   & 93.7  & 93.2  & 93.4  & -           \\
	& SEMv2~\citep{SEMv2}         & \textbf{97.50}   & 93.8  & 93.4  & 93.6  & {92.0}        \\
	& TRUST~\citep{Trust}         & 97.10         & -     & -     & -     & -           \\
	& SEMv3~\citep{SEMv3}         & \textbf{97.50}         & 94.8  & 95.4  & \textbf{95.1}  & 93.2        \\
	\midrule
	\multirow{4}{*}{\parbox{2cm}{Image-to-Text}} & EDD~\citep{EDD}           & 89.90         & -     & -     & -     & -           \\
	& TableFromer~\citep{TableFormer}   & 96.75         & -     & -     & -     & -           \\
	& VAST~\citep{VAST}          & 97.23         & -     & -     & -     & -           \\
	& Ours     & \textbf{97.50}  & \textbf{95.6}  & {94.7}  & \textbf{95.1}  & \textbf{94.0}        \\
	\bottomrule
\end{tabular}
\end{table*}

\subsection{Results} 
In this section, we evaluate the effectiveness of UniTabNet from three different perspectives. More details are provided in the Appendix~\ref{app:results}.

\noindent\textbf{Results from Electronic Document.}
As shown in Table~\ref{tab:comparison_pubtables1m}, compared to Image-to-Text approaches, our method has achieved a new state-of-the-art level. Although the bottom-up method~\citep{Detr} performs better on the GriTS-Loc metric, this is due to their use of the bounding box of the content within the cell to adjust the predicted bounding box of the cell. As illustrated in Table~\ref{tab:comparison_with_sota}, UniTabNet also performs comparably to the current advanced methods on the PubTabNet dataset.

\noindent\textbf{Results from Complex Scenarios.}
As shown in Table~\ref{tab:comparison_with_sota}, to demonstrate the robustness of UniTabNet in visual scenarios, we conduct experiments on the WTW and iFLYTAB datasets. On the WTW dataset, our method exhibits high precision but lower recall, primarily constrained by the maximum decoding length of the model. Therefore, compared to other non-autoregressive methods (Bottom-up and Split-and-merge), it achieves lower recall but comparable overall F1 scores with current methods. On the iFLYTAB dataset, UniTabNet achieves a new state-of-the-art performance.

\noindent\textbf{Results from Descriptive Tables.}
To demonstrate the effectiveness of UniTabNet in addressing descriptive tables, as shown in Table~\ref{tab:ablation_study}, we compare UniTabNet with the previously state-of-the-art SEMv3~\citep{SEMv3} on the iFLYTAB-DP dataset. SEMv3 is a purely visual approach for reconstructing table structures. However, iFLYTAB-DP contains a large number of tables with descriptive cells, requiring the model to understand the textual information within to make accurate structural predictions. The comparison shows that UniTabNet significantly outperforms SEMv3 in this scenario.


\begin{figure*}[t]
	\includegraphics[width=1\linewidth]{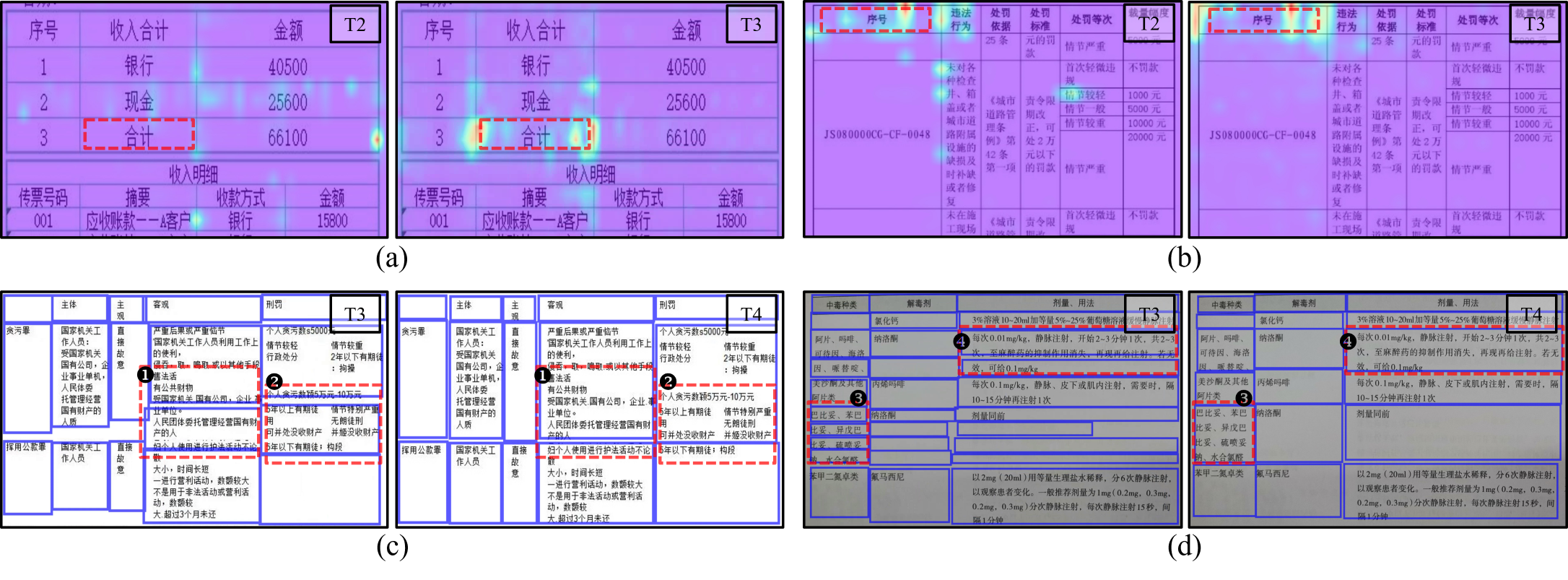}
	\caption{The illustration of the Vision Guider and Language Guider. Panels (a) and (b) compare the attention distributions within the decoding cells (regions indicated by red dashed boxes) for systems T2 and T3, respectively. Panels (c) and (d) display the comparative structural prediction results on iFLYTAB-DP for systems T3 and T4. The red dashed boxes highlight the regions where the predictions differ between the two systems, with system T4 accurately predicting in these areas.}
	\label{fig:05_ablation}
\end{figure*}

\subsection{Ablation Study}

\begin{table}[t]
	\centering
	\caption{Results of the TEDS-Struct evaluation for the UniTabNet model on the iFLYTAB and iFLYTAB-DB datasets. ``UL'' denotes ``Use of Uncertainty in Likelihood Optimization'' as detailed in Eq.~\ref{eq:ul_loss}. ``VG'' indicates the inclusion of a vision guider, and ``LG'' signifies the use of a language guider. ``D1'' and ``D2'' correspond to the performance metrics on the iFLYTAB validation set and iFLYTAB-DP set, respectively.}
	\label{tab:ablation_study}
		\begin{tabular}{cccccc}
			\toprule
			System & UL & VG & LG & D1 & D2 \\
			\midrule
			SEMv3 & - & - & - & 93.2 & 82.6 \\
			T1 & \xmark & \xmark & \xmark & 92.4 & 82.9 \\
			T2 & \cmark & \xmark & \xmark & 93.2 & 83.3 \\
			T3 & \cmark & \cmark & \xmark & 93.7 & 83.6 \\
			T4 & \cmark & \cmark & \cmark & 94.0 & 84.9 \\
			\bottomrule
		\end{tabular}
\end{table}

As shown in Table~\ref{tab:ablation_study}, to demonstrate the effectiveness of each module within the model, we design systems T1 through T4, which were evaluated on both iFLYTAB and iFLYTAB-DP datasets.

\noindent\textbf{The Effectiveness of Loss Design.}
During the entire training process of UniTabNet, the primary losses include regression loss and classification loss, which differ significantly in scale. Inspired by~\citep{WeightLoss}, we optimize the model by maximizing the Gaussian likelihood with homoscedastic uncertainty, as described in Eq.~\ref{eq:ul_loss}. Comparing systems T1 and T2 demonstrates the effectiveness of this loss design.

\noindent\textbf{The Effectiveness of Vision Guider.}
Table images are distinct from conventional document images, as each table cell provides unique visual cues linked to the corresponding row or column. In UniTabNet, we incorporate a Vision Guider at the final decoder layer to steer the model's focus towards pertinent visual segments of the table image. Figure~\ref{fig:05_ablation} illustrates the cross-attention mechanisms (averaged across the heads of the final layer) during the decoding stages of systems T2 and T3. The visualizations reveal that T3 more effectively concentrates on the regions pertaining to table cells throughout the decoding process. Furthermore, as shown in Table~\ref{tab:ablation_study}, T3 outperforms T2, demonstrating the effectiveness of the Vision Guider.

\noindent\textbf{The Effectiveness of Language Guider.}
Most previous methods for table structure recognition focus on reconstructing the table structure from a visual perspective. However, for tables rich in descriptive content, relying solely on visual cues can introduce ambiguities. In UniTabNet, we integrate a Language Guider into the final layer of the decoder, enhancing the model's capability to interpret the semantic content of the text. Figure~\ref{fig:05_ablation} displays the prediction results for systems T3 and T4 on the iFLYTAB-DP dataset, illustrating that T4 effectively mitigates visual ambiguities and improves text comprehension. Furthermore, as demonstrated in Table~\ref{tab:ablation_study}, T4 significantly outperforms T3 on the iFLYTAB-DP dataset, highlighting the effectiveness of the Language Guider.

\section{Conclusion}
In this paper, we present UniTabNet, a novel table structure recognition model leveraging the image-to-text paradigm, consisting of a vision encoder and a text decoder. UniTabNet employs a ``divide-and-conquer'' strategy to initially separate table cells, then uses physical and logical decoders to reconstruct cell polygon and span information. To improve visual focus and textual understanding within cells, we integrate a Vision Guider and a Language Guider in the text decoder. Comprehensive experiments conducted on publicly available datasets, including PubTables1M, PubTabNet, WTW, and iFLYTAB, demonstrate that UniTabNet achieves state-of-the-art performance in table structure recognition.

\section{Limitations}
Although UniTabNet has significantly streamlined the structure sequence of table outputs to only include two tokens: <C> and <NL>, its inference efficiency decreases as the number of table cells increases. Furthermore, due to limitations on maximum decoding length, UniTabNet exhibits relatively lower recall rates for table images with a large number of cells. Moreover, unlike the split-and-merge approach which utilizes a carefully designed merge module to handle a variety of table grid structures, UniTabNet employs classification to predict the span of rows and columns. This approach renders UniTabNet ineffective at dealing with previously unseen spans.

\bibliography{custom}
\clearpage
\appendix

\section{Appendix}
\subsection{Datasets and Evaluation Metrics}
\label{app:dataset_and_metric}
As shown in Table~\ref{tab:dataset_overview}, we summarize the datasets used during our experiments, along with the evaluation metrics employed to assess our model's performance on each dataset. We will detail each of these in the subsequent sections.

\noindent\textbf{PubTabNet.} PubTabNet is a large-scale table recognition dataset. PubTabNet annotates each table image with information about both the structure of table and the text content with position of each non-empty table cell. All tables are also axis-aligned and collected from scientific articles. The authors also proposed a new Tree-Edit-Distance-based Similarity (TEDS) metric for table recognition task, which can identify both table structure recognition and OCR errors. 	TEDS measures the similarity of the tree structure of tables. 
While using the TEDS metric, we need to present tables as a tree structure in the HTML format. 
Finally, TEDS between two trees is computed as:
\begin{align}
	\text{TEDS}(T_a, T_b) = 1 - \frac{\text{EditDist}(T_a, T_b)}{\max (\lvert T_a \rvert, \lvert T_b \rvert)}
\end{align}
where $T_a$ and $T_b$ are the tree structure of tables in the HTML formats. 
EditDist represents the tree-edit distance, and $\lvert T \rvert$ is the number of nodes in $T$. Since taking OCR errors into account may lead to an unfair comparison due to the different OCR models used by various TSR methods, we also employ a modified version of TEDS, called TEDS-Struct. The TEDS-Struct assesses the accuracy of table structure recognition, while disregarding the specific outcomes generated by OCR.

\begin{table*}[t]
	\centering
	\caption{The overview of datasets and respective metrics.}
	\label{tab:dataset_overview}
	\begin{tabular}{p{3cm}ccccccc}
		\toprule
		{Datasets} & \multicolumn{2}{c}{{Digital}} & \multicolumn{2}{c}{{Camera-captured}} & {Num} & {Metric} \\
		\cmidrule(lr){2-3} \cmidrule(lr){4-5}
		& {Wired} & {Wireless} & {Wired} & {Wireless} &              &                 \\
		\midrule
		PubTabNet~\citep{EDD}    & \cmark  & \cmark  & \xmark  & \xmark  & 568,000  & TEDS-Struct  \\
		PubTables1M~\citep{PubTables1M}  & \cmark  & \cmark  & \xmark  & \xmark  & 948,000  & GriTS        \\
		WTW~\citep{WTW}          & \cmark  & \xmark  & \cmark  & \xmark  & 14,581   & F1-Measure   \\
		iFLYTAB~\citep{SEMv2}      & \cmark  & \cmark  & \cmark  & \cmark  & 17,291   & TEDS-Struct  \\
		iFLYTAB-DP   & \xmark  & \cmark  & \xmark  & \cmark  & 322      & TEDS-Struct  \\
		\bottomrule
	\end{tabular}
\end{table*}

\begin{figure*}[t]
	\centering
	\includegraphics[width=1.\linewidth]{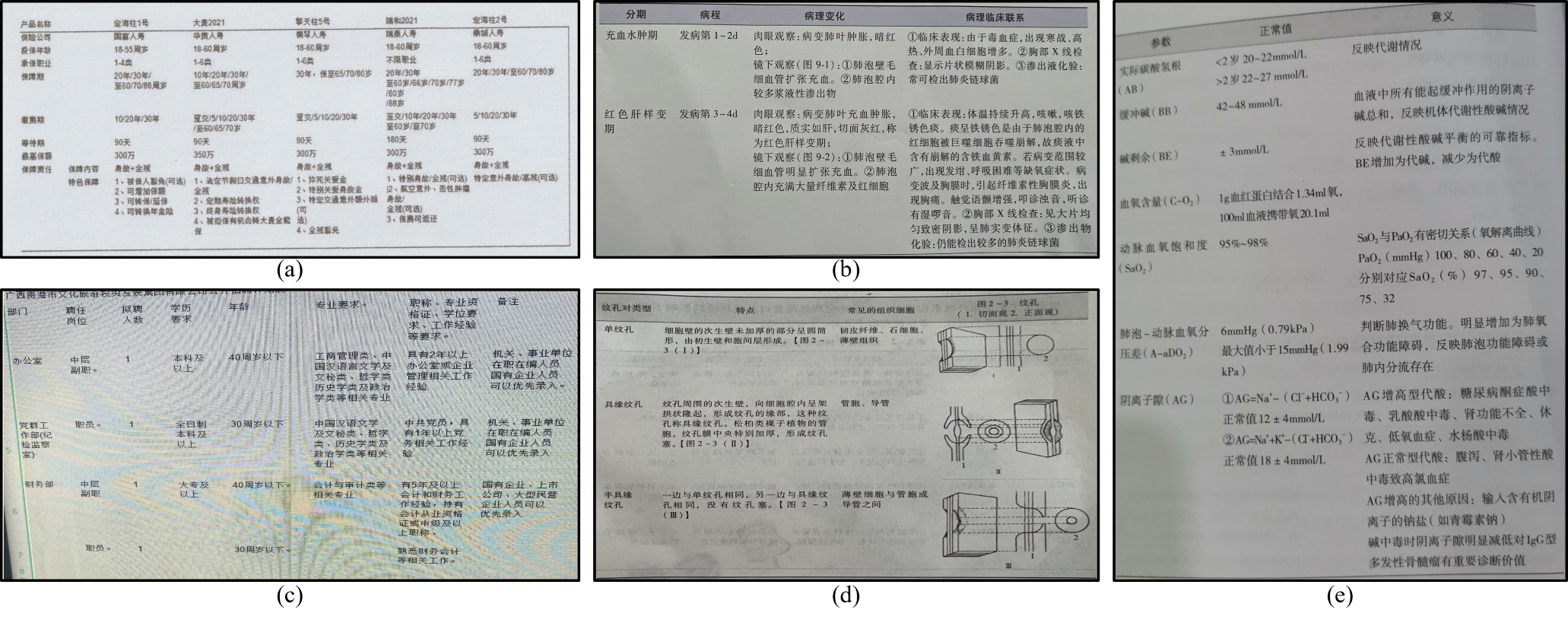}
	\caption{Some examples of the iFLYTAB-DP dataset.}
	\label{fig:01_app_iflytab_db}
\end{figure*}

\begin{figure*}[t]
	\centering
	\includegraphics[width=1.\linewidth]{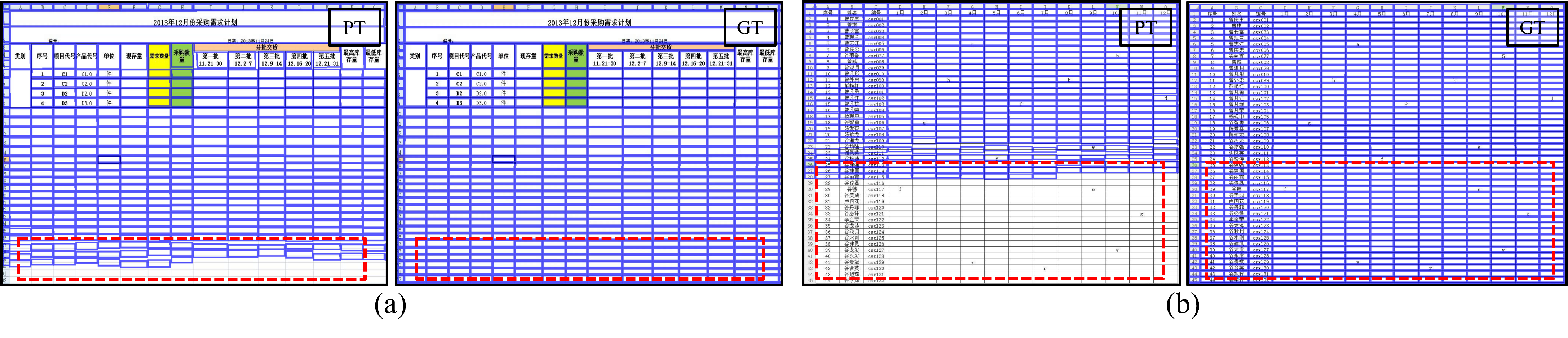}
	\caption{The illustration of the maximum decoding length limitation in UniTabNet. The samples are from the WTW dataset. The ``PT'' label in the top right corner of the image denotes the predicted results by UniTabNet, while ``GT'' indicates the ground truth structure of the table. Areas missed by the model due to the maximum decoding length limitation are highlighted with red dashed boxes.}
	\label{fig:02_app_lowrecall}
\end{figure*}

\begin{figure*}[t]
	\centering
	\includegraphics[width=1.\linewidth]{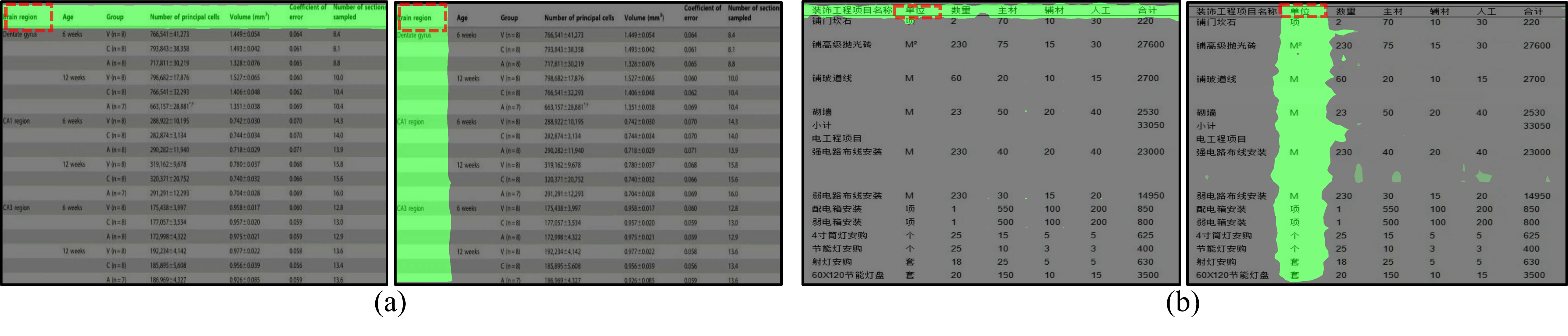}
	\caption{The illustration of row and column information learned by the Vision Guider. Panel (a) is from the PubTables1M dataset, and (b) is from the iFLYTAB dataset. The red dashed boxes highlight the area of the table cell currently being decoded. The green mask indicates the row and column information of the table cell as predicted by UniTabNet.}
	\label{fig:03_app_row_column}
\end{figure*}

\begin{figure*}[t]
	\centering
	\includegraphics[width=1.\linewidth]{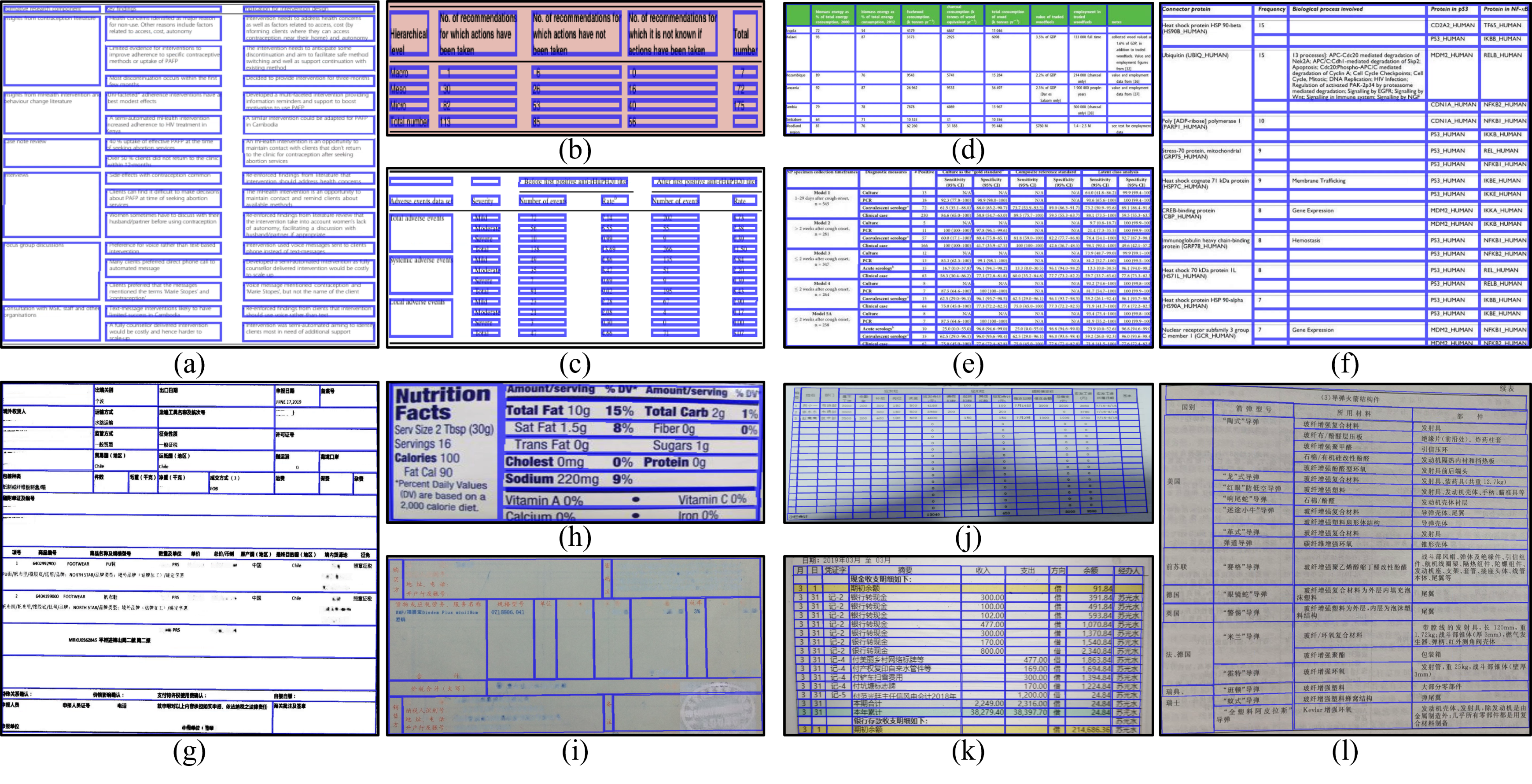}
	\caption{The prediction results of UniTabNet across different datasets. The blue boxes in the images represent the cell polygons decoded by UniTabNet. Panels (a) to (c) show predictions for the PubTabNet dataset, (d) to (f) for the PubTables1M dataset, (g) to (i) for the WTW dataset, and (j) to (l) for the iFLYTAB dataset.}
	\label{fig:04_app_results}
\end{figure*}

\noindent\textbf{PubTables1M.} Both the PubTables1M and PubTabNet datasets are sourced from the PubMed Central Open Access (PMCOA) database. The primary distinction between the two lies in the richness of annotation provided by PubTables1M. This dataset includes detailed annotations for projected row headers and bounding boxes for all rows, columns, and cells, encompassing even the blank cells. Additionally, it introduces a novel canonicalization procedure aimed at correcting oversegmentation. The purpose of this procedure is to ensure that each table is presented with a unique and unambiguous structural interpretation. To contrast our method with others, we evaluated it using the GriTS metric on this dataset. The recently proposed GriTS metric~\citep{Grits} directly compares predicted tables with the ground truth in matrix form and can be interpreted as an F-score reflecting the accuracy of predicted cells. Exact match accuracy is assessed by the percentage of tables for which all cells, including blank cells, are perfectly matched.

\noindent\textbf{WTW.} WTW dataset comprises 10,970 training images and 3,611 testing images, collected from wild and complex scenes. This dataset is specifically tailored to wired tabular objects and provides annotated information including tabular cell coordinates, and row/column data. We utilize the F1-Measure to evaluate our method on this dataset. To apply the F1-Measure, it is essential to detect the adjacency relationships among the table cells. The F1-Measure calculates the percentage of correctly detected pairs of adjacent cells, where both cells are accurately segmented and identified as neighbors. When evaluating on the WTW dataset, we employ the cell adjacency relationship metric~\cite{wtw-eval}, a variant of the F1-Measure. This metric aligns a ground truth cell with a predicted cell based on the Intersection over Union (IoU) criterion. For our assessments, we set the IoU threshold at 0.6.

\noindent\textbf{iFLYTAB.} 
The iFLYTAB dataset comprises 12,104 training samples and 5,187 testing samples. It offers comprehensive annotations for each table image, including physical coordinates and structural information. This dataset not only includes axis-aligned digital documents but also images captured by cameras, which present more challenges due to complex backgrounds and non-rigid image deformations. For evaluating our method on this dataset, we employ the official TEDS-Struct metric\footnote{\url{https://github.com/ZZR8066/SEMv2}\label{link:semv2}}. Specifically, during the evaluation process on iFLYTAB, we assign a distinctive marker to each text line, which signifies its individual content.

\noindent\textbf{iFLYTAB-DP.} To more precisely evaluate our model's performance on descriptive table images, we select 322 images from the iFLYTAB validation dataset, as shown in Figure~\ref{fig:01_app_iflytab_db}. To minimize the influence of visual cues such as table lines, which could assist the model's predictions, we specifically chose images of wireless tables. Our selection criteria primarily focuses on the presence of extensive textual descriptions within the cells. Additionally, we have contacted the authors of iFLYTAB, and they have agreed to make this subset of the dataset available on the official website soon$^{\ref{link:semv2}}$.

\subsection{Results}
\label{app:results}
In this section, we explain the issue of the relatively low recall rate exhibited by UniTabNet due to the limitation imposed by the maximum decoding length. As illustrated in Figure~\ref{fig:02_app_lowrecall}, we select some table images from the WTW dataset that contain a large number of cells. Due to the maximum decoding length constraint set at 500, this limitation significantly impacts the model's recall performance. However, as shown in Table~\ref{tab:comparison_with_sota}, UniTabNet achieves relatively high precision. When considering both precision and recall, UniTabNet's performance on the WTW dataset is comparable to current methods.

Additionally, as depicted in Figure~\ref{fig:03_app_row_column}, we visualize the row and column information learned by UniTabNet through the Vision Guider. The Vision Guider enables UniTabNet to focus more effectively on cell-related areas during the cell decoding process, as demonstrated in Figure~\ref{fig:05_ablation}.

Finally, Figure~\ref{fig:04_app_results} presents the prediction results of UniTabNet on the experimental datasets used. The model effectively processes both both simple and complex scenarios of table images. Notably, the cell polygons detected by UniTabNet in the PubTabNet dataset significantly differ from those in other datasets. This discrepancy arises because we directly use the official cell bounding box annotations provided, without any postprocessing.


\end{document}